\documentclass{article}

\usepackage[nonatbib,preprint]{neurips_2024}

\usepackage[utf8]{inputenc} 
\usepackage[T1]{fontenc}    
\usepackage{hyperref}       
\usepackage{url}            
\usepackage{booktabs}       
\usepackage{amsfonts}       
\usepackage{nicefrac}       
\usepackage{microtype}      
\usepackage{xcolor}         

\usepackage{natbib}
\setcitestyle{authoryear,open={(},close={)}} 

\usepackage{graphicx}
\usepackage{amsmath} 
\newenvironment{pm_boxed}
    {\begin{center}
    \begin{tabular}{|p{0.9\textwidth}|}
    \hline
    }
    { 
    \\\hline
    \end{tabular} 
    \end{center}
    }

\title{Multi-State-Action Tokenisation in Decision Transformers for Multi-Discrete Action Spaces}

\author{
    Perusha Moodley \\ School of Mathematical, Physical and \\ Computational Sciences,  University of Reading, \\ Reading, UK , \texttt{perusha.moodley@gmail.com}
    \And Pramod Kaushik \\ TCS Research \\ Pune, India \\ \texttt{ pramod.kaushik@tcs.com}
    \And Dhillu Thambi \\ Machine Learning Lab,  \\ IIIT Hyderabad, Hyderabad, India \\ \texttt{dhillu.thambi@research.iiit.ac.in}
    \And Mark Trovinger \\Department of Computer Science \\ and Engineering, Mississippi \\ State University,   Mississippi, USA \\ \texttt{mrt364@msstate.edu}
    \And Praveen Paruchuri \\ Machine Learning Lab,  \\ IIIT Hyderabad, Hyderabad, India \\ \texttt{praveen.p@iiit.ac.in}
    \And Xia Hong  \\ School of Mathematical, Physical and \\ Computational Sciences,  University of Reading, \\ Reading, UK , \texttt{x.hong@reading.ac.uk}
    \And Benjamin Rosman \\ School of Computer Science and Applied Mathematics,  \\ University of Witwatersrand, \\ Johannesburg, South Africa.  \texttt{benjros@gmail.com}
}

\begin{document}

\maketitle

\begin{abstract}

Decision Transformers, in their vanilla form, struggle to perform on image-based environments with multi-discrete action spaces. 
Although enhanced Decision Transformer architectures have been developed to improve performance, these methods have not specifically addressed this problem of multi-discrete action spaces which hampers existing Decision Transformer architectures from learning good representations. To mitigate this, we propose Multi-State Action Tokenisation (M-SAT), an approach for tokenising actions in multi-discrete action spaces that enhances the model's performance in such environments. Our approach involves two key changes: disentangling actions to the individual action level and tokenising the actions with auxiliary state information. These two key changes also improve individual action level interpretability and visibility within the attention layers.
We demonstrate the performance gains of M-SAT on challenging ViZDoom environments with multi-discrete action spaces and image-based state spaces, including the Deadly Corridor and My Way Home scenarios, where M-SAT outperforms the baseline Decision Transformer without any additional data or heavy computational overheads. Additionally, we find that removing positional encoding does not adversely affect M-SAT's performance and, in some cases, even improves it. 

\end{abstract}

\section{Introduction}
\label{sec:introduction}

Decision Transformers (DT) \citep{Chen2021-lk} are trained using reinforcement learning (RL) trajectories to generate actions in tasks of similar nature. Like other offline RL approaches \citep{Levine2020-th}, the Decision Transformer extracts behavioural policies from offline data without further interaction with the environment. The ability of transformers \citep{Vaswani2017-ju} to model long term dependencies and relationships in sequential data makes it particularly well suited to the offline RL problem.

In RL the nature of the task can impact the performance of these models, for example locomotion tasks versus navigation or goal-based tasks. Goal-based tasks, where the agent must achieve or navigate to a specific goal state represented in the environment, tend to be more challenging for Decision Transformers \citep{Emmons2021-yv}. RL agents operating in image-based environments with rich spatial structure, such as mazes or rooms, require the ability to reason about the relationships between states and actions. \citet{Emmons2021-yv} find that the standard Decision Transform struggles to capture and leverage this structural information effectively leading to suboptimal performance, proposing goal-state conditioning for multi-task, goal-conditioned environments such as Four-Room Gridworlds \citep{MinigridMiniworld23}, Pusher (Sawyer robot) \citep{Nair2018-sa} or AntMaze \citep{Fu2021-kt}, finding the original return-to-go conditioning performs better in locomotion-based environments without specific task goals.

Methods for enhancing transformer capabilities include developing advanced tokenisers \citep{Mathieu2023-nc, Reed2022-mq, Shang2021-zc}, using encoders to process multimodal encodings for transfer via cross attention \citep{Jaegle2021-tm, Jiang2022-pu}, new architectures \citep{Jaegle2021-tm, Jiang2022-pu, Mathieu2023-nc, Reed2022-mq, Jain2022-gl, Shang2021-zc, Wu2023-nx} and deriving special prompting methods \citep{Xu2022-kj, Jia2023-nk}. While these capabilities report promising results in a variety of environments including Mujoco \citep{Fu2021-kt} and Atari \citep{mnih2013playing}, results are mixed in more complex environments, often requiring more specifically handcrafted architectures sacrificing generality, and also making it inefficient needing larger volumes of data and more compute \citep{Mathieu2023-nc, Wang2023-oj, Reed2022-mq, Jiang2022-pu}.  

In this paper, we look at a specific class of action space called the multi-discrete action space in which RL agents must learn to select multiple discrete actions per timestep. An example would be in ViZDoom \citep{Kempka2016-eu} where the agent needs to both crouch and shoot as a parallel action in the same state. Current methods often convert multi-discrete actions to a single action representation \citep{Kanervisto2020-cu}, ignoring potential relationships between individual discrete actions thereby leading to sub-optimal performance in these datasets. On the other hand, human players are more likely to exploit these relationships when composing actions \citep{Dekker2022-ng}. In this paper we train Decision Transformers on expert data capable of extracting and utilising relational action insights in multi-discrete action space environments thereby leading to better performance in an efficient manner.
Ideally, when training Decision Transformer models, we want to capitalise on additional auxiliary information and structure in trajectory data to derive a more generalised model, without pre-training or use complex architectures. There is precedent for using auxiliary structure in transformer-models with other modalities \citep{Chen2021-fx, Wang2022-hs, Jaegle2021-tm} and auxiliary signals and structure are widely utilised in RL \citep{Jaderberg2016-iy, Zambaldi2018-lv}.   

We propose M-SAT (Multi-State Action Tokenisation), a method for capitalising on auxiliary action information in multi-discrete action spaces to improve the generalisation and model robustness of Decision Transformers in those spaces. 
Multi-discrete action spaces prove more difficult to train with Decision Transformer models so we propose tokenising actions at the individual action level with auxiliary state embedding information. The individual action token approach increases the visibility of these actions in the attention layers allowing the transformer to formulate relationships with, and determine the importance of, individual actions. Individual action tokens also facilitate the injection of auxiliary information at the action level and provides opportunities for action-level interpretability. Actions are tokenised with the preceding state embedding to encourage state-action associations and provide richer context to the action token. An additional impact on the model is increased stability during training and improved dynamics awareness. 
 
We test our approach on a set of challenging offline RL datasets generated from ViZDoom environments \citep{Kempka2016-eu} including Deadly Corridor and My Way Home, both featuring multi-discrete action spaces and image-based states. 
Our approach performs better than the baseline Decision Transformer in both environments. M-SAT improves Decision Transformer performance in Deadly Corridor and in My Way Home the agent is able to solve the environment while the baseline fails to do so. 
We run analysis and ablations to better understand model performance including removing the auxiliary state embeddings and find the state embedding is responsible for the improved performance of M-SAT. 
Given the navigation and goal-based nature of the environments, we test models with and without positional encoding. Here results are mixed but interesting, suggesting future avenues for research. 
We find the baseline's performance on My Way Home improves if we use a longer context length, suggesting that M-SAT is more sample efficient given that it solves the task with half the context.
The additional structure supplied by the state embedding results in a more robust model. The granular action tokenisation, aside from promoting the generalisation of the model, makes individual actions more visible and is more interpretable. The proposed solution does not require a pre-training phase or complex architecture and is able to exploit state and action structure in the environment with minimal modification to the Decision Transformer.

The specific contributions of this work include: \\
\vspace{1.0pt}
$\bullet$ Novel tokenisation of multi-discrete actions (M-SAT) in a Decision Transformer model that expands multi-discrete actions into individual actions before tokenising these actions with auxiliary state-embeddings. \\
\vspace{0.85pt}
$\bullet$ Demonstration of M-SAT on two ViZDoom multi-discrete environments yielding good performance on goal-based tasks and sample efficiency. \\
\vspace{0.85pt}
$\bullet$ Better interpretability due to disentangled action tokenisation \\
\vspace{0.85pt}

\section{Related Work}
\label{sec:related-work}
Incorporating auxiliary structure into transformer-based models has been approached in various ways. In language models, \citet{Chen2021-fx} intervene in a BERT-style model \citep{Devlin2018-xg} adding auxiliary token combination logic before classification, to improve performance on toxicity datasets. \citet{Qian2021-xb} explore incorporating structure into GPT-style models \citep{Brown2020-xc}, designing pre-training tasks that imbue the model with specific syntactic structure on limited size datasets. They find the resulting generative structural supervision produces a more robust model with human-like linguistic generalisation. 
\citet{Wang2022-hs} train an auxiliary module that captures relational dependencies between entities in an integrated energy systems setting and use this to inform the transformer model via cross-attention. 

A number of works deal with auxiliary information and structure at a more generalised level. \citet{Shi2023-kp} transfer structure from a language model to a Decision Transformer, finding the inherent reasoning abilities within language models prove useful when applied to RL tasks. \citet{Jaegle2021-tm, Jiang2022-pu, Reed2022-mq} target the generalist agent space which is a different but important problem space as generalist agents are often designed to handle auxiliary information and structure. Perceiver \citep{Jaegle2021-tm} is a generic architecture designed to manage different, high dimensional modalities such as video, audio and images with a single architecture. It encodes arbitrary input modalities into a compact latent bottleneck transferred via cross-attention to the transformer for iterative processing. This decoupling process allows the transformer to have a different capacity from the encoder and allows Perceiver to handle various data types while scaling linearly in compute, compensating for the lack of specific structure in the architecture by tagging inputs with position and modality-specific features. In complex RL robotic environments, VIMA \citep{Jiang2022-pu} uses extra modalities as auxiliary information when training tasks. Like Perceiver, VIMA alternates cross-attention of multimodal input data with self-attention layers. Gato \citep{Reed2022-mq} is designed to process multimodal data to cater for vastly different scenarios, including control tasks. The model is a study in how to tokenise and merge auxiliary information and supporting structure. 

While there is much work in the multimodal token space, fewer works focus on tokenisation related to Decision Transformers. 
StARformer \citep{Shang2021-zc} targets visual, image based RL environments, adopting a different tokenisation mechanism for training RL data. They first tokenise the timestep by compressing the state, action and reward information using a transformer encoder. The Decision Transformer is then trained over the compressed timestep token and a second tokenisation of the state, using rewards for prompting. Relating to the importance of tokens in transformer models, FNET, \citep{Lee-Thorp2021-io} investigates token mixing mechanisms, replacing the self attention mechanism with a simpler linear transformation that mixes tokens. They find FNET achieves a significantly high percentage of accuracy compared with BERT models \citep{Devlin2018-xg}, highlighting that one of the key functions of the attention mechanism is token mixing. Their finding supports the expansion of action tokens in our work to promote token mixing. 

The Graph Transformer model of \citet{li2023what} uses Laplacian Positional Encodings to inject graph structural information into the node representations. These dataset based Laplacian position encodings, derived from the eigenvectors of the graph Laplacian, are a generalisation of the sinusoidal position encodings used in the original Transformer, allowing Graph Transformers to capture important positional and structural relationships in graph data.

Finally, while Decision Transformers have been used on environments with multi-discrete actions, the examples observed to date have not focused on the action space in particular \citep{Mathieu2023-nc, Reed2022-mq}.

Two papers that propose approaches similar to those in this paper were subsequently discovered by the authors and are mentioned below. 
\citet{Boige2023-xn} investigate several methods when pre-training transformer agents over offline datasets, one of which is tokenisation at the component level for actions and states finding this results in a more generalised model. While findings are similar, the nature of our investigations are different, as is the nature of the environment and tasks we tackle. They consider non-image based environments and focus on effective methods for robust pre-training of transformers. 
\citet{fujimoto2023for} introduce a state-action dynamics representation in the TD7 architecture, trained using a dynamics-based loss. Unlike our paper their key focus is online RL with some offline tests, however they target non-image state spaces. 

\section{Preliminaries}
\subsection{Offline reinforcement learning with  multi-discrete action}
Let the reinforcement learning (RL) in a Markov decision process (MDP) be described by the tuple $(\mathcal{S}, \mathcal{A}, P, R)$. In the setting of the multi-discrete action, at time step $t$, the MDP tuple consists of states, $s_t \in \mathcal{S}$, multi-discrete actions $\mathbf{a}_t \in \mathcal{A}$, transition dynamics $P(s_{t+1} |s_t, \mathbf{a}_t)$  and a reward
function $r_t = R(s_t, \mathbf{a}_t)$, where $\mathbf{a}_t =  [a_t^{1}, a_t^{2},   ..., a_t^{n}] $ with $a_t^{i}  \in \{0, 1, ..., N_i\}$  for $i = 1, 2, ..., n$. We also denote  a trajectory   composed of a sequence of states, actions, and rewards as $\tau = (s_0, \mathbf{a}_0 , r_0, s_1, \mathbf{a}_1 , r_1, . . . , s_T , \mathbf{a}_T , r_T )$. 
The goal in offline reinforcement learning is to learn a policy which maximizes the expected return  $E\left[ \sum_{t=1}^{T} r_t\right]$. The offline reinforcement learning is restrictive, as its learning setting  removes the ability for agents to explore the environment and collect additional feedback.

\subsection{Decision Transformers}
\label{sec:background-decision-transformers}
The Decision Transformer \citep{Chen2021-lk} is a transformer \citep{Vaswani2017-ju} adapted from offline RL training over trajectory data generated by an RL algorithm or collected manually from humans. While in the offline RL setting \citep{Levine2020-th} an agent is trained with no interaction with the environment during training,  the Decision Transformer is able to extract behavioural policies from sequential trajectory data similar to language modelling for sequential texts. Let the return-to-go (RTG) reward of a trajectory at time step  $t$ be  $\hat{R}_t =\sum_{i=t}^{T}r_i$, which is the expected sum of future rewards from a given state, calculated for each trajectory step by accumulating the returns from that step forward to the end of the trajectory or context.

Note that conventional reinforcement learning is a special case of multi-discrete action by reducing the number of concurrent actions $n$ to one. The trajectory representation of the Decision Transformer \citep{Chen2021-lk} is given by 
$$\tau =\left (\hat{R_1} , s_1, a_1, \hat{R_2} , s_2, a_2,..., \hat{R_T} , s_T, a_T \right),$$ 
with $T$ timesteps of state, action, return tuples, spanning complete or partial episodes. $\hat{R_t}$ is used in preference to the every step reward, $r_t$, as a prompt, which allows the transformer to generate actions based on future desired returns.

In the Decision Transformer, each modality (return, state, or action) is passed into an embedding network (convolutional encoder for images, linear layer for continuous states). The embeddings are then processed by an auto-regressive transformer model, trained to predict the next action given the previous tokens using a linear output layer. Tokenisers are networks that encode each mode in the timestep to the model dimension, $d_{model}$, a property of the transformer, and are selected to match the type of data, so for example, a convolutional neural network (CNN) could be used for image states. A multi-layer perceptron (MLP) is used for discrete tokenisation instead of the action embedding, so MLPs are used to tokenise the discrete and continuous values typically associated with actions and RTGs. As per the original transformer \citep{Vaswani2017-ju}, the timestep or position is also encoded and added to each embedded mode. The final batch of input data after tokenisation has dimensions batch size$\times$context length$\times$model dimension, where batch size, model dimension and context length are hyperparameters. The Decision Transformer receives a context of trajectory data and uses this prompt to generate the next action. 
Since there are environments where relationships between actions exist \citep{fan2022minedojo, Kempka2016-eu, DBLP:journals/corr/abs-1708-04782, Jain2022-gl, Chitnis2020Intrinsic}, which could be exploited to produce more intuitive and performant agents, here we consider the Decision Transformer with multi-discrete actions, and the trajectory representation becomes
$$\tau = \left( \hat{R_1} , s_1, \mathbf{a}_1, \hat{R_2} , s_2, \mathbf{a}_2,..., \hat{R_T} , s_T, \mathbf{a}_T \right).$$

\section{Proposed M-SAT approach using structured tokenisation}
\label{sec:approach}

We propose Multi-State-Action Tokenisation (M-SAT), an extension to the Decision Transformer that augments multi-discrete action tokenisation to exploit action-action and state-action relationships. This is achieved by first exposing individual actions during tokenisation and second, by injecting auxiliary state-related structure into the token during training. The Decision Transformer architecture remains largely unchanged, and there is no pre-training involved.

\subsection{Tokenisation and action generation}
\label{sec:approach-tokenisation}
\begin{figure}[h]
  \centering
  \fbox{\includegraphics[width=0.65\linewidth]{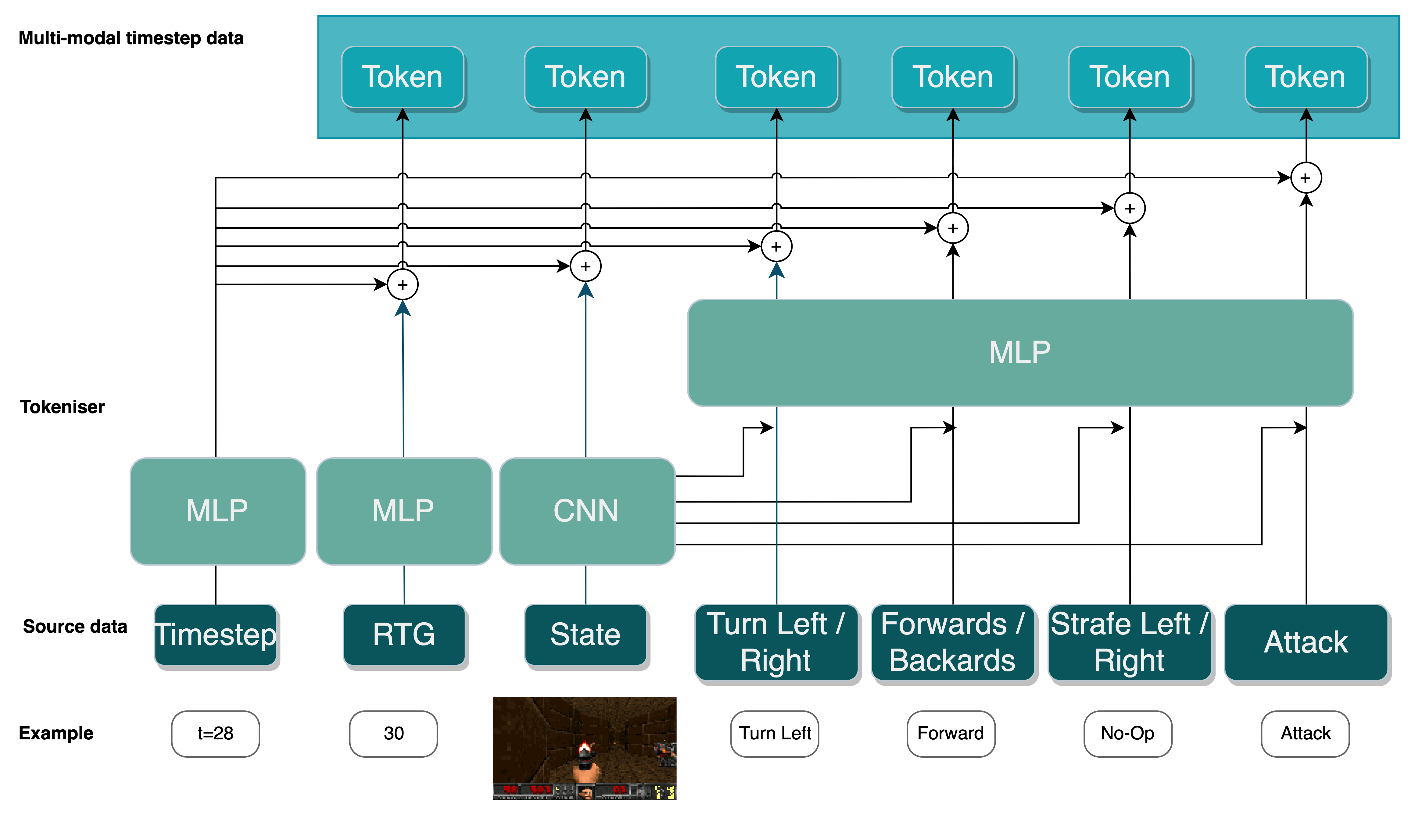}}
  \caption{Tokenisation of the RTG, state and multi-discrete actions in the M-SAT Decision Transformer, with the proposed multi-token actions}
  \label{fig:tokenisation}
\end{figure}

The original Decision Transformer \citep{Chen2021-lk} processes multimodal data including RTG, state and action into a single token per mode. As a result, an $N$-dimensional multi-discrete action would be tokenised into a single token. We propose a change to the way multi-discrete actions are tokenised for the Decision Transformer, illustrated in Figure \ref{fig:tokenisation}. 
M-SAT features a novel action tokeniser that:
\begin{itemize}
    \item Expands multi-discrete actions to individual component actions 
    \item Encodes each action with the preceding state embedding as auxiliary information, generating a single token per individual action of model dimension $d_{model}$.
\end{itemize} 

More concretely, a multi-discrete action with $N$ discrete action components is separated into $N$ one-hot encoded vectors. Each one-hot encoded action is concatenated with the state embedding vector for the timestep and is tokenised by an MLP, generating $N$ action tokens of $d_{model}$ for $N$ discrete actions. Decision Transformers enhanced with this tokenisation approach are referred to as M-SAT. The remainder of the action generation process is unmodified; multi-discrete actions are generated by sampling from categorical distributions derived from state logits, as per the original Decision Transformer. The pseudo-code for this process is provided in Appendix \ref{sec:appendix-pseudocode}.  

The tokenisation process effectively encodes state or environment features with each action. This is analogous to positional encoding that tags each token with a timestep, but is conceptually closer to the dataset-derived positional encoding in \citet{li2023what}. The motivation for the token expansion is to increase the visibility of individual actions in the attention layers, making more efficient use of the attention capabilities of transformers. The attention layers will see $N$-tokens per multi-discrete action instead of one token, allowing more relationships to form between tokens.  
The state-action tokens are proposed to convey auxiliary structural hints to the attention blocks. The intuition is the attention heads should have maximum visibility of all information, structural and relational. In M-SAT the attention heads of the Transformer will see multiple action tokens per timestep as part of the context, providing more opportunity for the heads to also formulate relationships between individual actions. The causal nature of the underlying GPT model means current timesteps refer to historical tokens or context to learn which tokens to attend to when generating a new action. The process is auto-regressive with previously generated actions appended to the context when generating new actions. In M-SAT individual actions are now more directly informing future actions. We intimate the state-action connection is important for solving goal-based tasks in image-based environments, which is the focus of our study. 

\section{Experiments}

\subsection{Experimental Setup}
\label{sec:exp-setup}
We evaluate M-SAT in goal-based environments based in labyrinths and rooms, featuring image-based state spaces and multi-discrete action spaces. 
The ViZDoom environment \citep{Kempka2016-eu} features several scenarios with multi-discrete action spaces including Deadly Corridor (DC) and My Way Home (MWH). In DC the task requires the agent to walk down a corridor populated at intervals by attacking adversarial entities to reach a goal at the end (see Figure \ref{fig:deadly_corridor}). It features a multi-discrete action space with four discrete actions selected on each timestep (Table \ref{tab:deadly-actions} - Left), for example, an action with Left, Forward, No-Op and No-Op is represented as $[1,1,0,0]$.  

\begin{table}[h]
    \centering
    \caption{ViZDoom multi-discrete action spaces for Deadly-Corridor and My Way Home. At each timestep, the agent must jointly select one action from every row.}
    \label{tab:deadly-actions}
    \begin{tabular}{c c}
        Deadly Corridor & My Way Home  \\
    \begin{tabular}{|c |c |c |c |}\hline
         &0 & 1 & 2  \\ \hline
        $a^1$: &No-Op & Left & Right  \\ \hline
        $a^2$: &No-Op & Forward & Backward \\ \hline
        $a^3$: &No-Op & Strafe Left & Strafe Right \\ \hline
        $a^4$: &No-Op & Attack & - \\ \hline
    \end{tabular} & 
        \begin{tabular}{|c |c |c |c |}\hline
         &0 & 1 & 2  \\ \hline
    $a^1$: &No-Op&Turn Left &Turn Right \\ \hline
    $a^2$: &No-Op&Forward &Backward \\ \hline
    \end{tabular}

    \end{tabular}

\end{table}

In MWH the task is to navigate a labyrinth of interconnected rooms and corridors to reach a goal (green vest) within a time constrained period. The reward is obtained once the goal is achieved. Unlike DC, the agent is randomly spawned in any location, adding an element of complexity. The action space is outlined in Table \ref{tab:deadly-actions} - Right. 

In the ViZDoom environment, the state is composed of two components: a first-person perspective image with dimensions $3 \times 72 \times 128$ and an environment feature vector that includes the agent's health. In this work, only the image component was utilised for training the Decision Transformer.

\begin{figure}[htbp]
    \centering
    \includegraphics[width=0.9\linewidth]{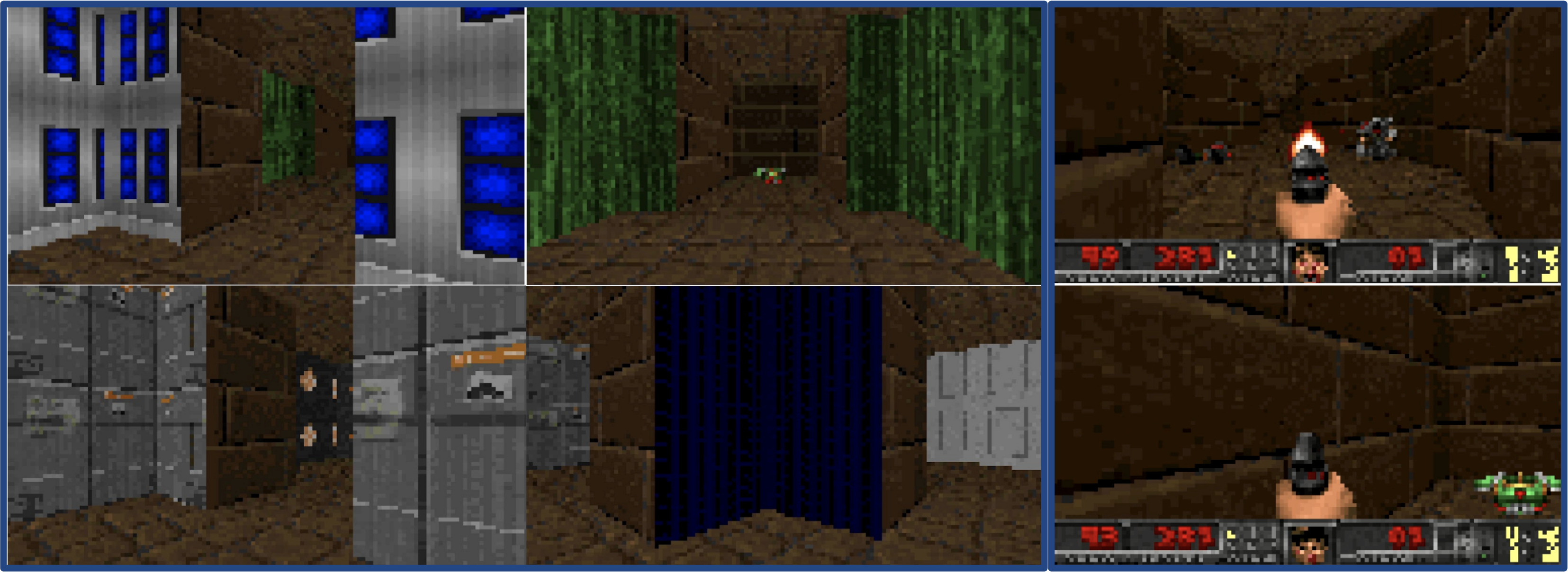}
    \caption{Sample of image-based states from My Way Home (Left) with multiple rooms and corridors leading to a green vest goal state and Deadly Corridor (Right) with attacking adversaries and a green vest goal state}
    \label{fig:vizdoom_states}
\end{figure}

\subsubsection{Baseline Method}
\label{subsec:exp-baseline}
The baseline approach uses the original Decision Transformer code \citep{Chen2021-lk}, modified to process multi-discrete actions to a single action token using an MLP, encoding a $1\times 4$ action for Deadly Corridor and a $1\times 2$ action for My Way Home to $1\times d_{model}$ where $d_{model}=128$. Further implementation details are provided in the Appendix \ref{sec:appendix-dt-implementation-details}.

\subsubsection{Multi-Action Tokens (M-SAT) }
\label{subsec:exp-m-sat}
In M-SAT the dimension of the multi-discrete action determines how many action tokens are generated per timestep. 
This results in 4 actions per timestep in DC and 2 for MWH of dimension $d_{model}$. The overall impact is 3 extra tokens per timestep for DC and 1 extra token for MWH. This impacts context length, for example in DC, this would convert a context length of $60$ to a context of $360 = 60 \times 6$. 
The original Decision Transformer code was modified to perform the additional action  tokenisation inside the model and to cater for the changing context lengths.

\subsection{Methods of Analysis}  
\label{sec:approach-analysis}
The comparative and ablative methods to M-SAT  were designed to address several key questions:  
\begin{enumerate}   
\item Does a Decision Transformer trained with multiple action tokens outperform single action tokens in a multi-discrete action scenario?  
\item In the multiple action token configuration, are actions tokenised with auxiliary state embeddings more effective?  
\item Does removing all position encoding improve the multiple action token model?  
\end{enumerate}
\subsubsection{No Position Encoding: } \label{subsec:variant-pe}
M-SAT expands multi-discrete actions to multiple tokens, increasing token information available to the Decision Transformer in multi-discrete environments. Position encoding (PE) is how the transformer is informed about the sequencing or position of a token, viewed as providing relevant structural information to the attention mechanism. We investigate the impact of PE by removing all PE tying states, actions, and RTGs to a timestep, to improve the visibility of the newly proposed individual action tokens in attention layers. \citet{Kazemnejad2023-ao} found that removing PE in language models resulted in the network learning a form of relative encoding to replace it. Following this, we removed PE to further increase token mixing opportunities between all tokens and trigger a PE more suitable for RL environments.

\subsubsection{Multiple Action Tokens (MAT)}
This experiment investigates the importance of the state-action tokenisation in multi-discrete environments. M-SAT tokenises actions with state embeddings in a method reminiscent of positional encoding. We question whether auxiliary information when tokenising actions is necessary and propose MAT (Multiple Action Tokens). MAT simply tokenises multi-discrete actions into individual discrete action tokens, without state-action associations, to determine their informational value.

\subsection{Results}
\label{sec:results}

\begin{figure}[htbp]
        \centering
    \begin{minipage}[c]{1.0\textwidth}
    \begin{minipage}[c]{0.5\textwidth}
        \includegraphics[width=0.9\linewidth]{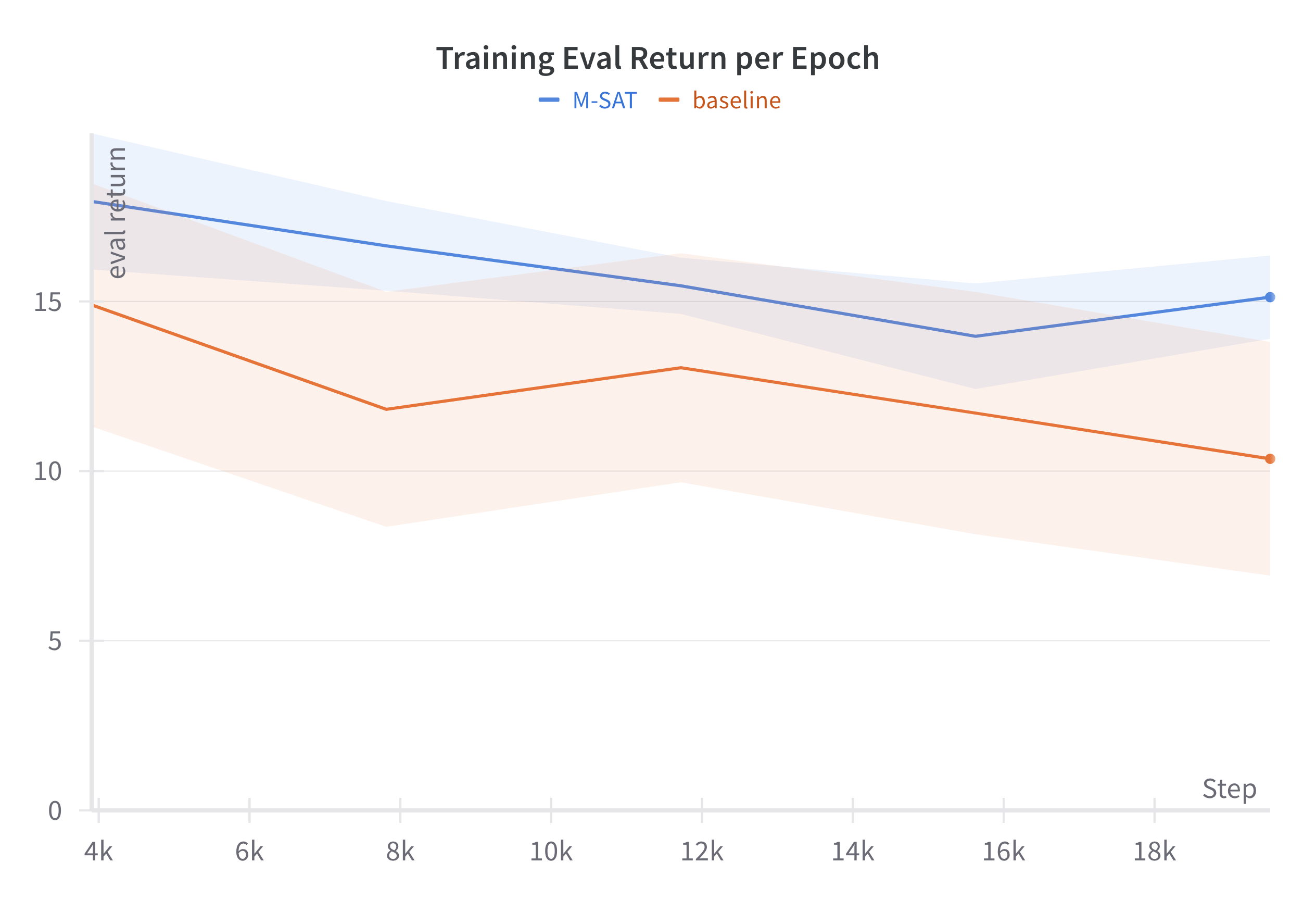}
    \end{minipage} %
    \begin{minipage}[c]{0.5\textwidth}        
        \includegraphics[width=0.9\linewidth]{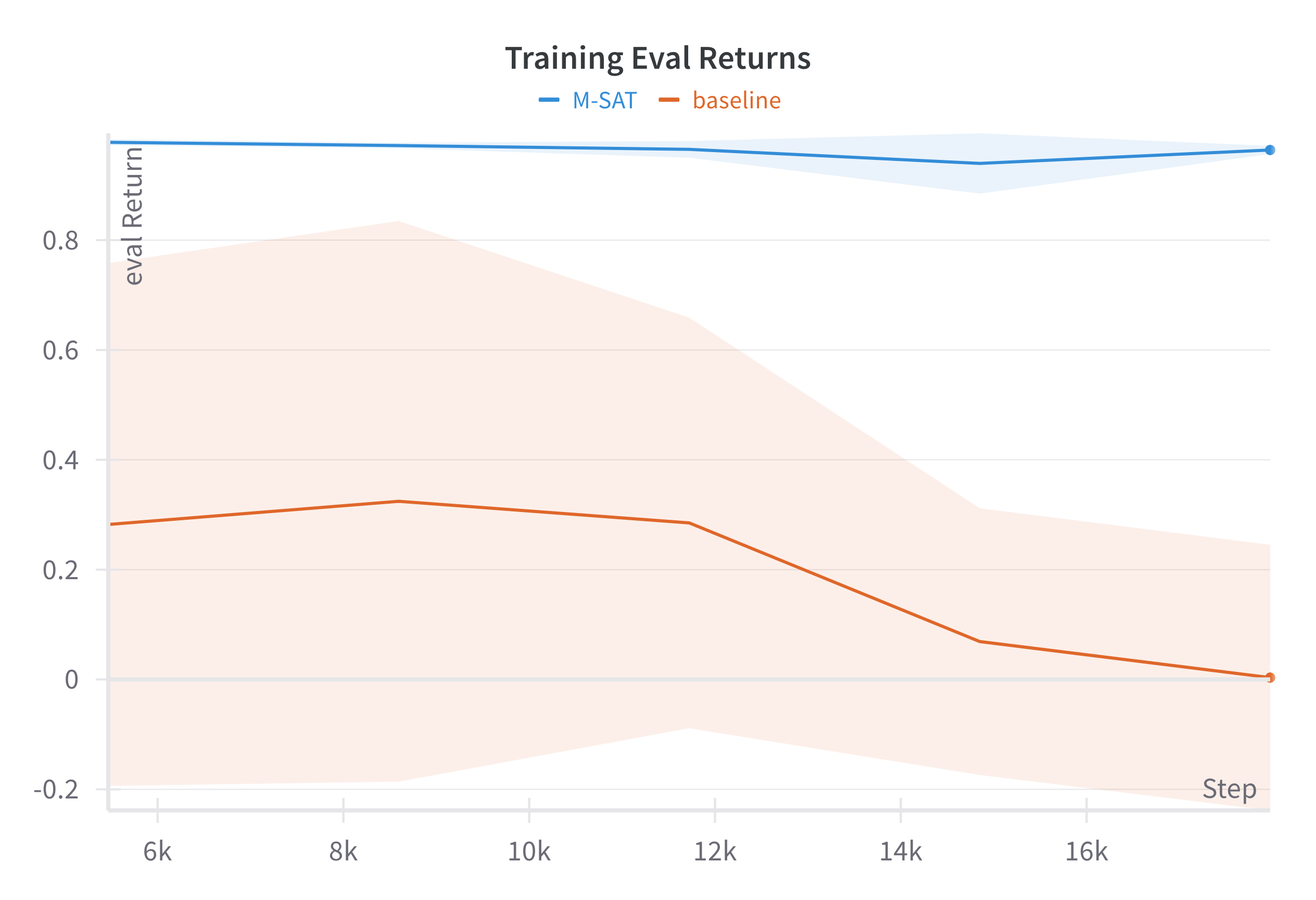}
    \end{minipage} 
    \end{minipage}
    \caption{Evaluation results over 50 runs and 5 seeds for DC (left) and MWH (right). See Appendix \ref{sec:appendix-dt-implementation-details} for training details}
    \label{fig:results}
\end{figure}

Figure \ref{fig:results} shows the evaluation results for the baseline and M-SAT models after each epoch. M-SAT achieves a higher result on average, also demonstrating lower variance than the baseline model. In DC M-SAT consistently attains better results than the baseline. In MWH the baseline fails to solve the task on average while M-SAT achieves an almost perfect score. This is arguably a harder environment with the agent randomly spawned in any room, facing any direction. In both environments the M-SAT scores demonstrate a better learnt representation of the dynamics of the environment. The lower variances implies more model stability, provided by the state-action grounding. Results support the proposed design of M-SAT, to expand multi-discrete actions to multiple action tokens to defer structure extraction and relationship development to the attention mechanism. In the baseline the multi-discrete action is compressed into a single token by the tokeniser (MLP), so the tokeniser determines which actions are important prior to the attention mechanism. Our hypothesis is that the attention mechanism should determine individual action relevance instead of the tokeniser. This modification is designed to exploit the powerful relational capabilities of transformer models, making tokens available at a granular level for information mixing within attention layers. In M-SAT, individual actions are encoded by an MLP and received by the attention layer as individual tokens, allowing the attention heads to form state-action, action-action or RTG-action relations. In the baseline model, the single action tokenising MLP effectively encodes action relationships when generating the token, providing less information for attention block to formulate relationships by comparison. Results suggest deferring complex processing, such as relationships, to attention heads rather than tokenisers.

Table \ref{tab:mat-results-1} provides results for the ablations, specifically for M-SAT with and without PE and for ablating M-SAT without auxiliary state information, namely MAT. 
In both environments, M-SAT outperforms MAT, suggesting the state embedding is contributing useful information to the action tokenisation. MAT performs better than the baseline on average but in MWH the difference between MAT and the baseline is significant, implying extra action tokens can make a positive difference to solving tasks in multi-discrete action spaces.
The No PE ablation was motivated, particularly in the case of MWH with random spawning of the initial state, by removing constraints that hampered the recognition of previously seen states. With PE, the model must learn to recognise the same state in different positional encodings (positions or timesteps), which requires more time or data to train. Removing the PE means states are more easily recognised and the model can learn faster. No PE in M-SAT produces different results in DC and MWH, with DC showing an overall improvement but not MWH. This may be related to the difference in nature of the environments. In DC there is more similarity between corridor states which would benefit from removing PE (refer to Figure \ref{fig:deadly_corridor} in Appendix \ref{sec:appendix-env-details}). The state-action token from M-SAT benefits from this too given the repetitive nature of the task, encountering three monsters at intervals along a corridor. There is less similarity within the MWH task by comparison. Removing PE can be helpful in some circumstances however there is insufficient data to make a generalisation and this is left for future work. 

\label{sec:results-ablations}
\begin{table}[ht]
    \centering
    \caption{Table of Results for MWH (Left), DC (Right) comparing M-SAT with and without PE and MAT, over 100 evaluation runs and 5 seeds. Maximum reward for MWH is 1.0 and 22.0 for DC}
    \label{tab:mat-results-1}
    \begin{tabular}{c c}
    \begin{tabular}{|c |c |c |}
    \hline
    Model Type & Ave Eval Rewards \\
    \hline
    M-SAT No PE & $0.863 \pm 0.256$  \\
    \hline
    M-SAT with PE& $0.964 \pm 0.005$ \\
    \hline
    MAT & $0.747 \pm 0.249$  \\
    \hline
    \end{tabular}     
         &  
    \begin{tabular}{|c |c |c |}
    \hline
    Model Type & Ave Eval Rewards \\
    \hline
    M-SAT No PE & $16.17 \pm 0.33$  \\
    \hline
    M-SAT with PE& $14.88 \pm 0.44$ \\
    \hline
    MAT & $12.28 \pm 3.1$  \\
    \hline
    \end{tabular}     
         
    \end{tabular}
\end{table}

The state-action associations in M-SAT outperforms MAT, suggesting the attention heads have detected useful structure in the state-action token. In M-SAT, states are pre-pended to discrete actions before tokenisation in a method reminiscent of dataset-derived position encoding. There is precedent for using dataset-specific positional encoding. \citet{li2023what} derive a Laplacian PE from a graph dataset to allow the Graph Transformer to capture and leverage the positional information that is relevant to graph-related tasks. We may think of the image states encoded with the actions as playing a similar role in goal-based tasks. Rather than a sequential PE we are using a spatial PE that may be more relevant in some RL environments.

\subsubsection{Interpreting the model}

In Mechanistic Interpretation \citep{Elhage2021-yd, bricken2023monosemanticity, sharma2023understanding} trained models are reverse engineered to understand the algorithms and representations learnt. While tools have been developed for large language transformer models, there are currently fewer options for understanding the behaviour of Decision Transformers. Using tools like Torchlens \citep{Taylor2023} to extract attention scores during evaluation allows us to inspect the attention layers and is useful for identifying behaviour patterns. The trained M-SAT models for DC and MWH were evaluated in this way and CircuitsViz \citep{circuitvis} was used to visualise attention at the token level for the generated evaluation trajectory. 

Figure \ref{fig:token-heatmap-1} in Appendix \ref{app:mi} demonstrates the attention heads focusing on the individual action tokens in the DC environment. 
The empirical data suggests a state where the agent attacks influences the attack action $a^4$ (refer to Table \ref{tab:deadly-actions}) in several attention heads, implying the multi-token action representation allows for a granular level of influence that the single-token action representation will not be able to experience or reveal. 
In Figure \ref{fig:mwh-l6-actions}, Appendix \ref{app:mi}, several heads in layer 6 focus on the action tokens and display action-related behaviours. The figure illustrates action-specific behaviour where the agent navigates a corner using a sequence of actions requiring a similar pattern for turning. The timesteps all refer back to a set of turning actions. This repeats throughout the episode with different action patterns visible in the heads. 
Appendix \ref{app:mi} provides details on other attention visualisations.
The M-SAT architecture allows for a better interpretable model compared to the traditional Decision Transformer for multi-discrete actions.

\section{Conclusion}
\label{sec:conclusion}
In this paper we proposed M-SAT, a method for tokenising multi-discrete actions to multiple action tokens with auxiliary state information in Decision Transformers. Experiments conducted in the ViZDoom multi-discrete environments, Deadly Corridor and My Way Home, demonstrated that the proposed M-SAT method outperformed other approaches. 
A comparison of results with a baseline single action token model and ablative methods demonstrated that multiple action token models achieve higher scores with lower variance overall in both environments. In My Way Home the baseline was unable to solve the task while M-SAT achieved close to optimal performance with a shorter context length compared to the baseline. A further benefit of the proposed model was the improved interpretability opportunities; with all actions exposed, attention heads were analysed at the individual action level. The results are limited to goal-based tasks in sparsely rewarded, multi-discrete environments and further work is required to generalise to additional settings and to non-image-based state spaces. 
\paragraph{Future Work:} More studies on the impact of positional encoding on M-SAT is recommended. Results from the PE ablations were different for the two environments which bears further investigation. Additional methods in Mechanistic interpretation for RL settings should be explored to improve the understanding of agent behaviour. We recommend exploring more complex environments and action spaces in future.

\begin{ack}
We are grateful to and acknowledge the Farama Foundation who provided and maintain ViZDoom Wrappers. We have cited and also acknowledge the use of the original Decision Transformer code, Sample Factory, TorchLens and CircuitViz.  
\end{ack}

\bibliography{references}

\section{Appendix / supplemental material}

\subsection{M-SAT tokenisation - Pseudocode}
\label{sec:appendix-pseudocode}

\begin{pm_boxed}
    \textbf{Pseudo-code for M-SAT action tokenisation} \\
    \hline
    \begin{flushleft}
\textbf{Inputs}: States with dimension $(B,T, n_{channels},height,width)$, \\
Actions with dimension $(B, T, n_{actions})$ \\
Where $n_{actions}$ is the number of elements in the multi-discrete action, \\
B is the batch size and T is the context length \\
$d_{model}$ is the model dimension \\
$d_{action}$ is the dimension of the multi-discrete action \\
$(n_{channels},height,width)$ are the dimensions of the state images \\
\textbf{Method}: Action Tokenisation \\
\begin{enumerate}
    \item Tokenise state through CNN to get state tokens $(B, T, d_{model})$
    \item One-hot encode the actions individually such that the new action dimensions are $(B, T, n_{actions}, d_{action})$ and each individual action has effective dimension $(B, T, d_{action})$
    \item Concatenate the state token $(B, T, d_{model})$ to each one-hot encoded action $(B, T, d_{action})$; the resultant dimension is $(B, T, d_{model} + d_{action})$
    \item Tokenise the concatenated state-action batch through the State-Action embedding MLP to get action tokens of dimension $(B, T, n_{actions}, d_{model})$
    
\end{enumerate}

    \end{flushleft}
\end{pm_boxed}

\subsection{Environment Details}
\label{sec:appendix-env-details}
\begin{figure}[htbp]
    \centering
    \includegraphics[width=0.8\linewidth]{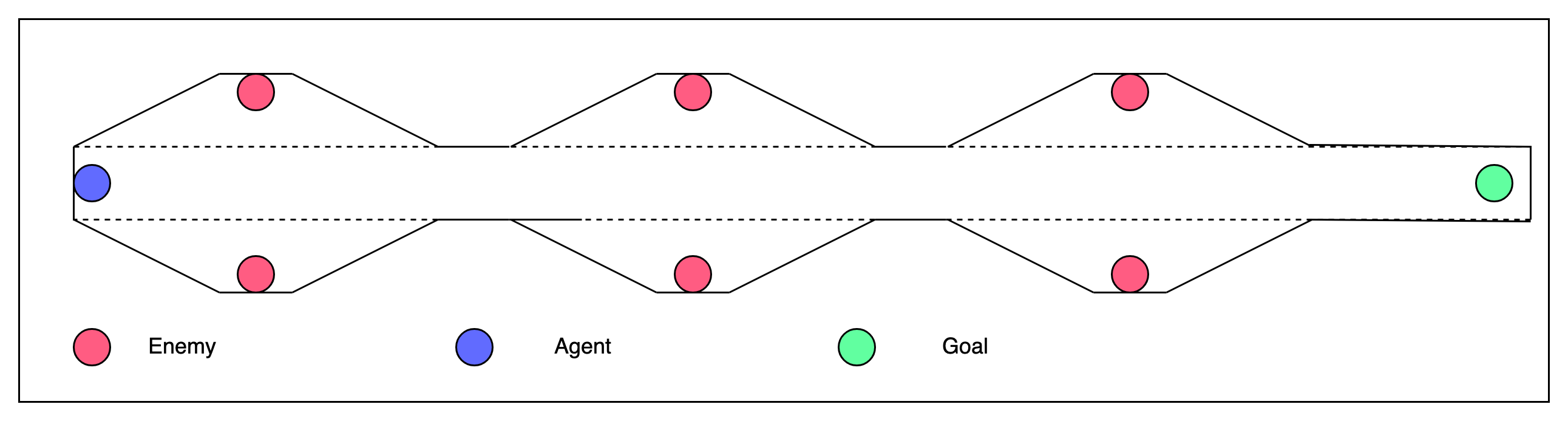}
    \caption{Simple map outline of ViZDoom's Deadly Corridor scenario showing the locations of the agent, enemies and the goal. }
    \label{fig:deadly_corridor}
\end{figure}

The ViZDoom environment \citep{Kempka2016-eu} was designed for use in machine learning modelling initiatives including RL. The environment is configurable, with multiple scenarios with varying layouts, action spaces and reward functions possible. The environment is challenging for RL algorithms with features including a state space comprising colour images from a first person perspective, random spawning of monsters that attack the agent, the random spawning of the agent in maze-type scenes and limitations on health or agent lifespan. There are scenarios with multi-discrete action spaces, two of which are used in this paper including Deadly Corridor and My Way Home.  
The ViZDoom  Deadly Corridor (DC) scenario has a 4 dimensional multi-discrete action space and involves the agent walking down a corridor, populated by attacking monsters at fixed locations, to reach the goal (a green vest), at the end as illustrated in Figure \ref{fig:deadly_corridor}. 
The My Way Home (MWH) scenario is more maze-like, where the agent must successfully navigate several rooms and corridor to eventually locate the goal (again a green vest). The agent is randomly spawned in any location and facing any direction at the start of the episode. There are no attacking elements however there is a timelimit. The action space is a 2-dimensional multi-discrete action. 

\subsubsection{States in ViZDoom}
The ViZDoom environment provides a state with two components, viz. a first-person perspective image (dimensions $3\times72\times128$) and an environment feature vector with the agent's health and other information, depending on the scenario. In this work, only the image was used as the state when training the Decision Transformer. Some commonly used image pre-processing techniques derived by \citet{mnih2013playing} for pixel-based image states in Atari were applied to the ViZDoom images, including resizing and frame skipping. A frame skip of 4 was adopted as per \citet{Kempka2016-eu} who demonstrated that higher skip-rates learnt faster in general but suffered with poor fine-grained control. Also in keeping with \citet{Kempka2016-eu}, colour images were used rather than the gray-scaling often performed on Atari images.

\subsubsection{Rewards in ViZDoom}

\paragraph{Deadly Corridor}
A per step reward is calculated, based on the current coordinate of the agent, with the agent starting at coordinate 0, and the end of the tunnel is coordinate ~1300.
If the agent reaches the goal, a green vest at the end of the corridor, it receives an additional +1000 reward (on top of the ``x-coordinate" reward).
There is no specific incentive for attacking monsters, however there is a death penalty of -100 if the agent dies. Finally, there is a time limit of 2100 timesteps to reach the end of the corridor before the episode terminates. \\

\paragraph{My Way Home} 
This scenario consists of a number of rooms connected by corridors. The goal is a green vest located in one of the rooms and the agent is randomly spawned in a room at the start of the episode. The episode ends when the vest is reached, with a reward of +1 or on timeout. There is an every step penalty of -0.0001. 

\subsection{Decision Transformer Related Setup and Implementation}
\label{sec:appendix-dt-implementation-details}
A few aspects pertaining to the Decision Transformer are discussed below, including the context length and further details related to tokenisation. 

\paragraph{Training specification} 
All models were trained on a server with 40 CPU cores, 200GB RAM and an Nvidia GPU (RTX A5000 with 24GB RAM). Training times for each epoch were between 20-40 minutes. 

\paragraph{Context Length}
The context length is the number of timesteps the Decision Transformer sees during training, resulting in an input batch dimension of [batch size$\times$context length$\times$token dimension]. The context length is a hyperparameter that is optimised to suit the task. 
Initially a context length of 30 was adopted for both DC and MWH. While this context worked effectively for MWH the same was not true for DC. After tuning a context length of 60 was adopted. 

DC: 
\cite{Chen2021-lk} comment that the context length can be important for enabling credit assignment and mention that the Key-to-Door environment was more complex, requiring a context of at least the length of a full episode. Similarly, \cite{Laskin2022-ot} found that in-context learning in RL was more likely when the context length covered a full episode, adding that the context often needed to span at least 2-4 times the episode length. 
These findings influenced the selection of a context length for DC by providing sufficient context for the transformer to see the attainment of the goal at the end of the trajectory within the context. Based on these findings a context length of 60 was selected after analysing the offline trajectory data and ensuring sufficient episode trajectories were available where the agent reached the end of the corridor and obtained a reward. 

MWH: 
The context length of 30 proved sufficient for MWH. The context length was increased to 60 to determine whether longer contexts could improve the baseline performance that was very poor by comparison to M-SAT. Results in Table \ref{tab:mat-results-2}.

\label{sec:results-mwh}
\begin{table}[ht]
    \centering
    \caption{Table of Results for My Way Home, Context Length 60: Baseline, MAT and M-SAT are compared over 100 evaluation runs and multiple seeds, with and without PE. Maximum reward for MWH is 1.0}
    \label{tab:mat-results-2}
    \begin{tabular}{|c |c |c |}
    \hline
    Model Type & Ave Eval Rewards \\
    \hline
    Baseline & $0.243 \pm 0.178$   \\
    \hline
    M-SAT No PE & $0.932 \pm 0.186$  \\
    \hline
    M-SAT with PE& $0.928 \pm 0.172$ \\
    \hline
    MAT No PE& $0.956 \pm 0.101$ \\
    \hline
    MAT with PE& $0.960 \pm 0.092$ \\
    \hline
    \end{tabular} 

\end{table}

The baseline's performance is significantly improved. MAT now outperforms M-SAT and the PE loses significance in both MAT and M-SAT. 
The implication is with a sufficiently long context, the Decision Transformer can learn to solve complex tasks. Techniques such as those used in M-SAT can make the training process itself more efficient in terms of computation and memory usage, with convergence appearing much sooner. 
Interestingly M-SAT with context 30 still achieves the highest performance overall for MWH. 

\subsection{Decision Transformer Hyperparameters}
\label{sec:dt-hyperparameters}
Details of the key hyperparameters for the Decision Transformer are provided in Tables \ref{tab:dt-hyperparams-mwh} and \ref{tab:dt-hyperparams-dc} below.  
These hyperparameters were used for both the single-action-token baseline and multi-action-token models.
 
\begin{table}[ht]
    \centering
      \caption{Decision Transformer Hyperparameters for training My Way Home - expert dataset}
    \label{tab:dt-hyperparams-mwh}
    \begin{tabular}{|l|c|}\hline
        env & My Way Home\\
        layers & 8 \\
        heads & 8 \\
        number of timesteps of training data & 500k \\
        epochs & 5 \\
        embed dropout & 0.1 \\
        residual dropout & 0.01 \\
        attention dropout & 0.1 \\
        batch size & 128 \\
        learning rate & 0.005 \\
        warmup & 10240 \\
        context length & 30 \\
        eval target & 1.8 \\ \hline
    \end{tabular}
  
\end{table}

\begin{table}[ht]
    \centering
      \caption{Decision Transformer Hyperparameters for training Deadly Corridor - expert dataset}
    \label{tab:dt-hyperparams-dc}
    \begin{tabular}{|l|c|}\hline
        env & Deadly Corridor\\
        layers & 8 \\
        heads & 8 \\
        number of timesteps of training data & 500k \\
        epochs & 5 \\
        embed dropout & 0.1 \\
        residual dropout & 0.1 \\
        attention dropout & 0.3 \\
        batch size & 128 \\
        learning rate & 0.005 \\
        warmup & 2000 \\
        context length & 60 \\
        eval target & 29 \\ \hline
    \end{tabular}
  
\end{table}

\subsection{Data Generation} \label{sec:6-approach-data-generation}

\begin{figure}[htbp]
    \centering
    \includegraphics[width=1.0\linewidth]{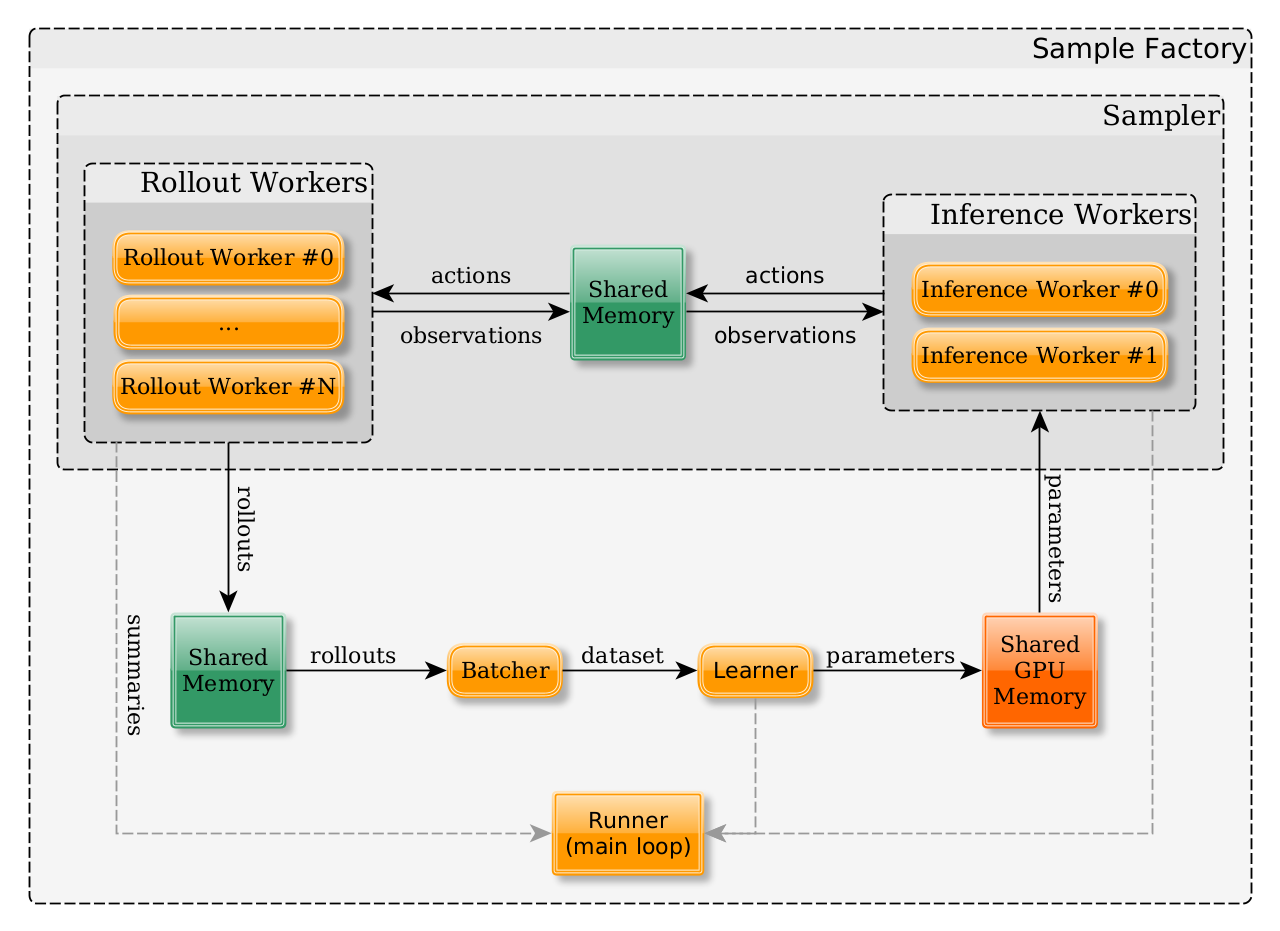}
    \caption{Sample Factory architecture overview - Source: \citep{petrenko2020sf}}
    \label{fig:sf-architecture}
\end{figure}

\begin{figure}[htbp]
        \centering
        \includegraphics[width=0.6\linewidth]{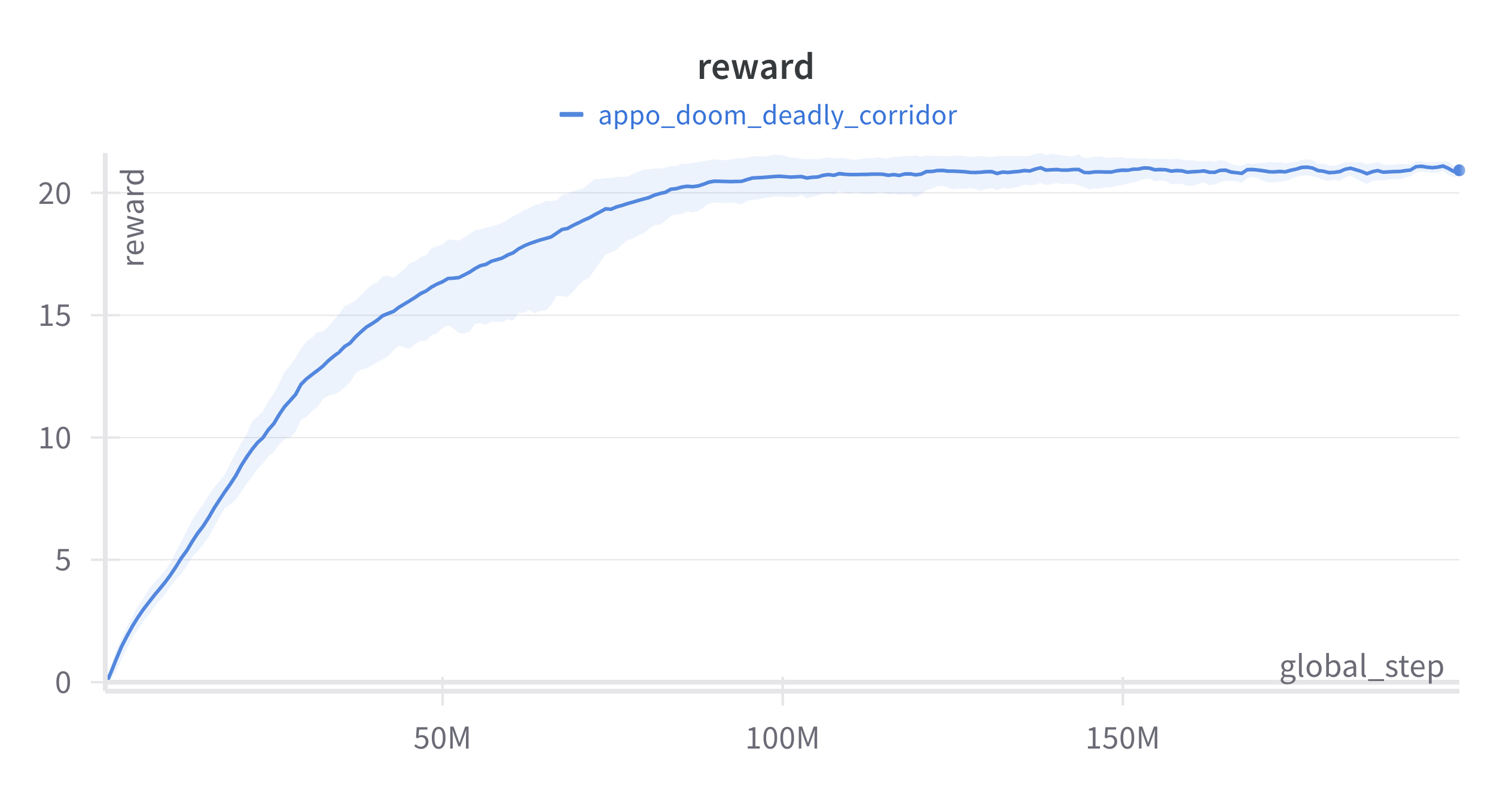}
        \caption{APPO Training curves for Deadly Corridor training - reward (Sample Factory)}
        \label{fig:dc-sfplots-reward}
\end{figure}

\begin{figure}[htbp]
        \centering
        \includegraphics[width=0.6\linewidth]{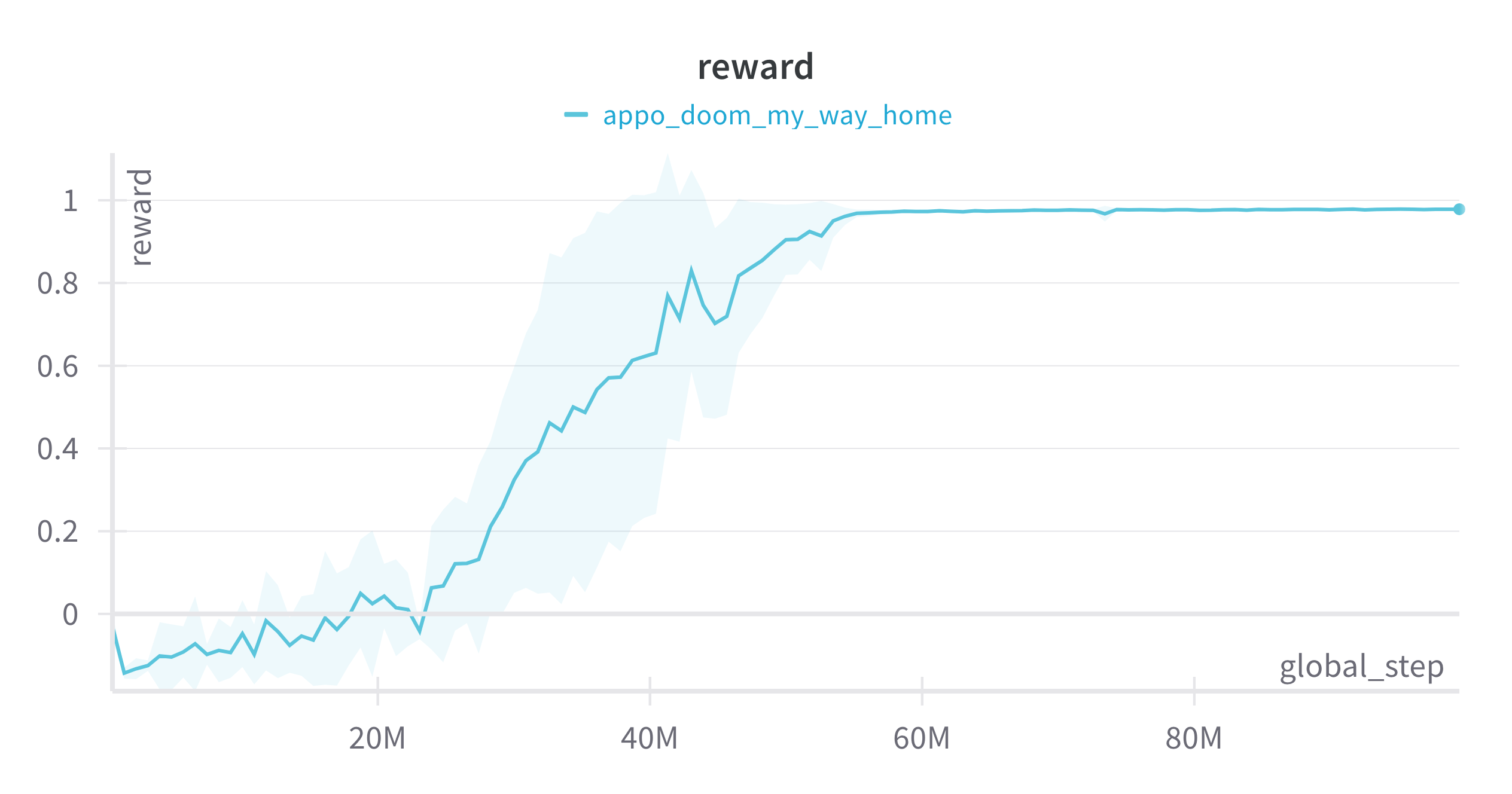}
        \caption{APPO Training curves for My Way Home training - reward (Sample Factory)}
        \label{fig:mwh-sfplots-reward}
\end{figure}

The Asynchronous-PPO algorithm (APPO) \citep{petrenko2020sf} was used to generate offline data for training the Decision Transformer. The APPO algorithm uses a combination of parallelisation of training components, including multiple instances of the environment and multiple workers for rollouts and inference, as illustrated in Figure \ref{fig:sf-architecture}. For the DC scenario the APPO agent was trained for 200 million frames until reward evaluations stabilised (optimal or close to optimal/expert), after which approximately 1 million frames of data was collected. The training curves for APPO are provided in Figures \ref{fig:dc-sfplots-reward} and \ref{fig:mwh-sfplots-reward} for DC and MWH respectively. 

Table \ref{tabel4} below provides the key hyperparameters required for generating the offline data in SF. The data was generated on a server with 40 CPU cores, 200GB RAM and an Nvidia GPU (RTX A5000 with 24GB RAM). These specifications were used to design the hyperparameters and achieve optimal performance from the APPO algorithm, as recommended by SF. 

\begin{table}[hb]
\centering
\caption{Relevant command line arguments to replicate data collection in Sample Factory}\label{tabel4}
\begin{tabular}{|l|c|} \hline
      env&doom-deadly-corridor \\
        num-workers&80 \\
        num-envs-per-worker&8 \\
        batch-size&4096 \\
        train-for-env-steps&200000000 \\
        with-wandb& True \\
        rollout&128 \\
        use-rnn&True \\ \hline
\end{tabular}
\end{table}

\clearpage

\subsection{Interpretability Plots}
\label{app:mi}

\begin{figure}[htbp]
    \centering
    \includegraphics[width=1.1\linewidth]{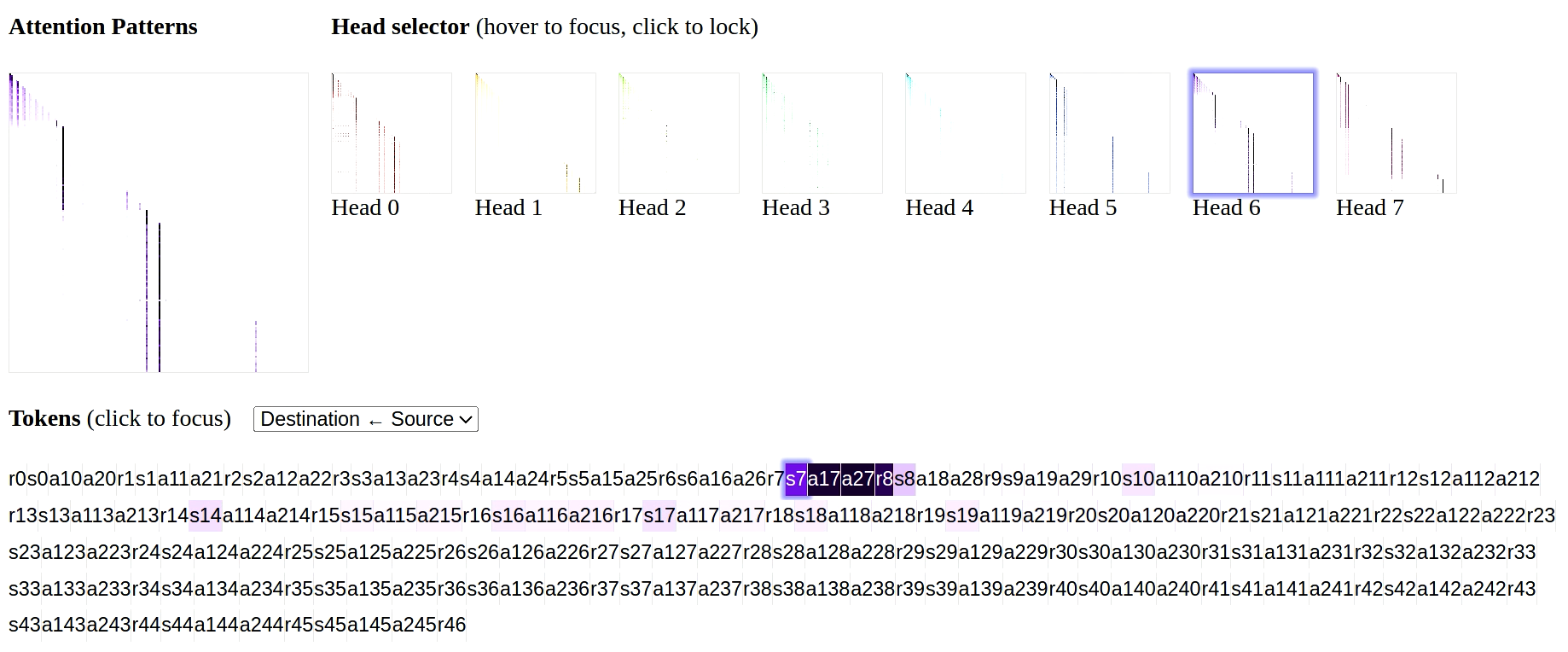}
    \caption{A sample plot of an attention layer from the Decision Transformer using Circuitviz, a tool for visualising attention at the token level. }
    \label{fig:circuitvis}
\end{figure}

Circuitviz \citep{circuitvis} was used to interrogate the trained Decision Transformer. A sample of the html file generated is illustrated in Figure \ref{fig:circuitvis} where the attention heads are displayed in the top section and the sequence of tokens in the bottom section. The html file is interactive so hovering over a token highlights the influence it has on other tokens. Tokens for timestep $t$ are labelled as r$t$ for RTGs, s$t$ for states and a$it$ where $i$ is the action index within the multi-discrete action. In MWH the multi-discrete action has 2 components so timestep 17 would be labelled a$117$ and a$217$. 

\begin{figure}[h]
    \centering
    \includegraphics[width=1.0\linewidth]{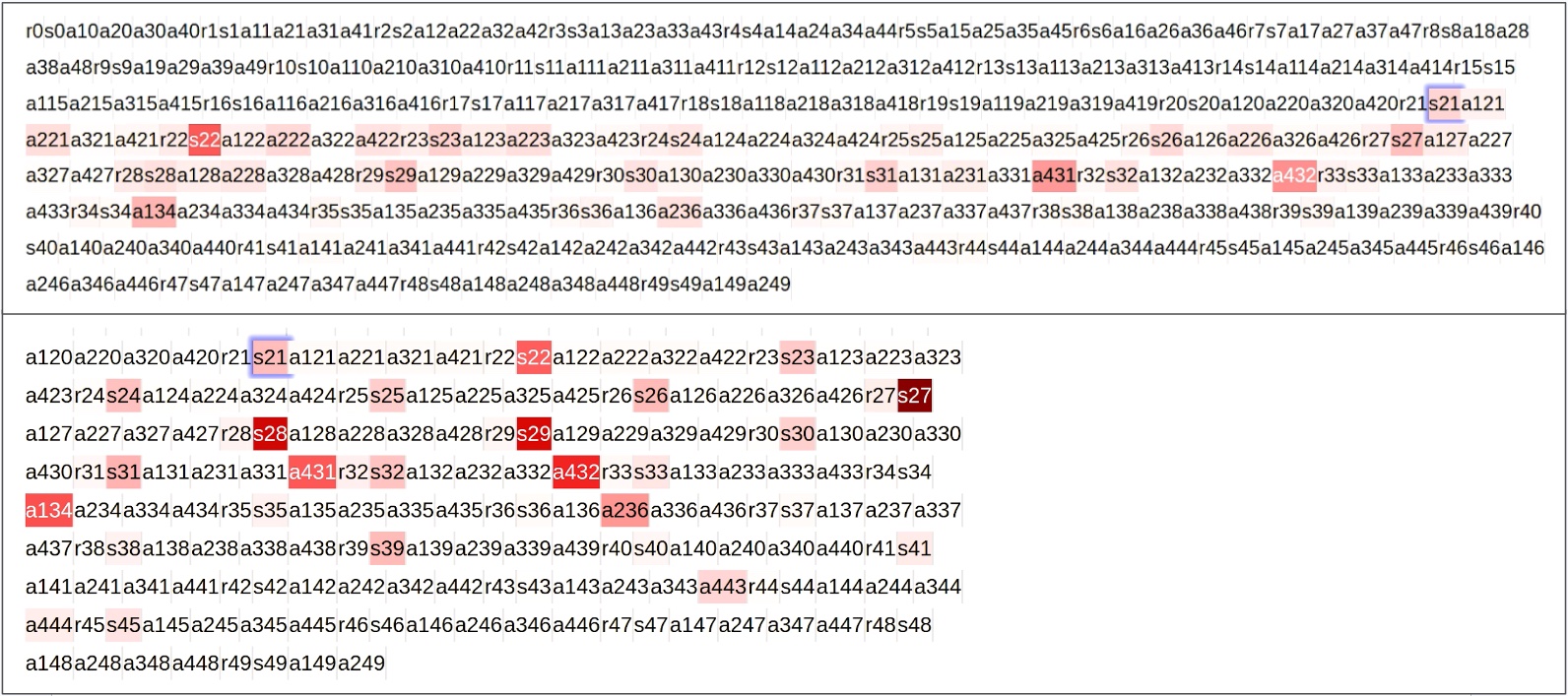}
    \caption{The top snippet is from Attention layer 1, head 0 and the bottom snippet from layer 2, head 0 - both show the influence of state s21 on future tokens, including individual action tokens. Refer to Figure \ref{fig:vizdoom-game-states} to view a sample of the associated states from this evaluation trajectory. }
    \label{fig:token-heatmap-1}
\end{figure}

\begin{figure}[h]
    \centering
    \includegraphics[width=1.0\linewidth]{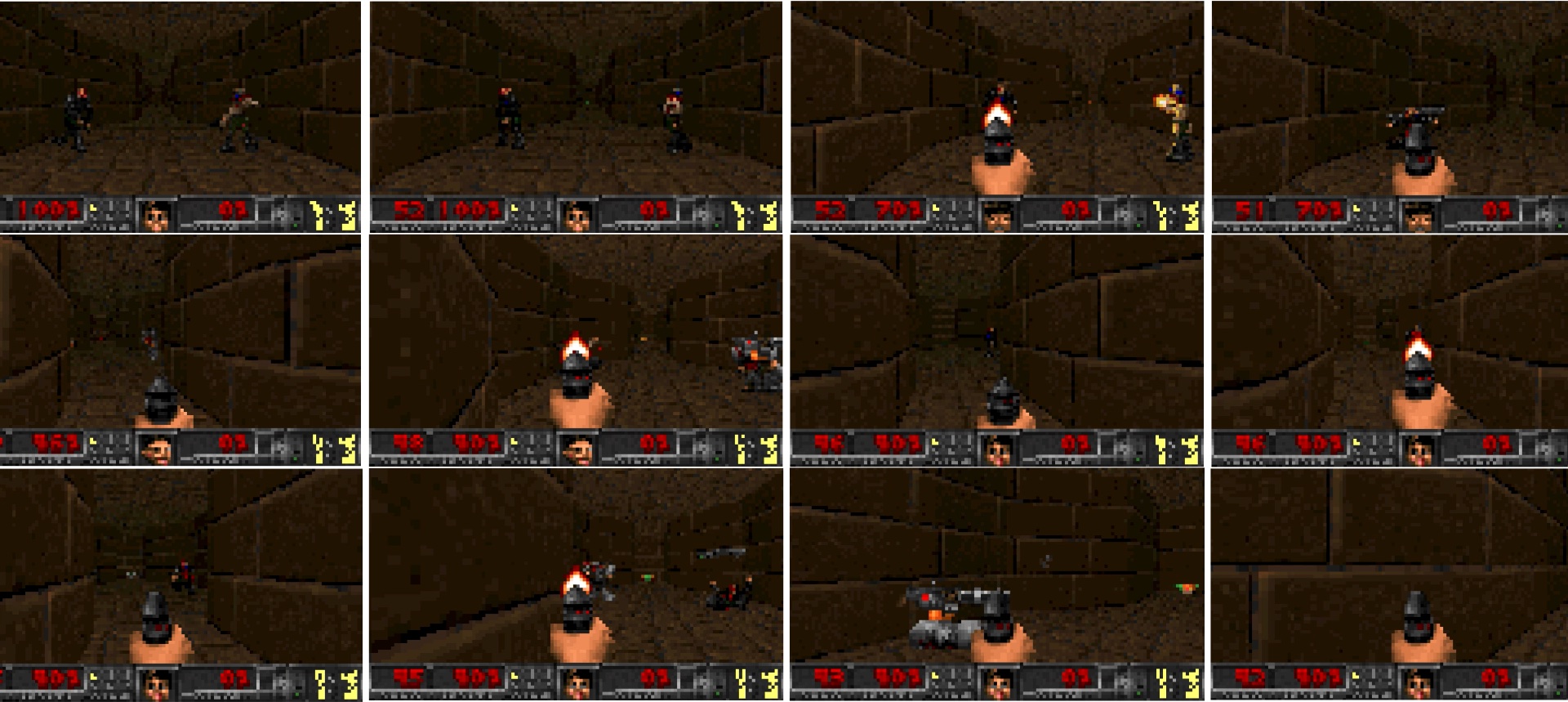}
    \caption{states from ViZDoom dataset, supporting attention heatmap snippets (Figure \ref{fig:token-heatmap-1}) from left to right, top to bottom: S1, S2, S5, S7, S16, S21, S31, S32, S33, S35, S40, S44}
    \label{fig:vizdoom-game-states}
\end{figure}

With this basic visualisation approach we first consider the trained DC model. 
Figure \ref{fig:token-heatmap-1} exhibits the attention from two heads, layer 1 head 0 and layer 3 head 0. These segments show that these heads view state s21 as important by allowing it to influence both states and actions that follow. Figure \ref{fig:vizdoom-game-states} shows state s21 to be firing upon an enemy. The heatmaps imply that some information from this state is transmitted to future states and also to three action tokens, notably the a4 (attack action position) tokens for timesteps 31, 32. States 32 and 33 show the impact of those actions, with the enemy coming into range again and the agent firing. The empirical data suggests the firing state s21 influences the firing action a4 directly in at least two attention heads, implying the multi-token action representation allows for a granular level of influence that the single-token action representation will not be able to experience or reveal. This pattern is repeated in several heads but the figure displays a sample to demonstrate the concept. 

\begin{figure}[h]
    \centering
    \includegraphics[width=1.0\linewidth]{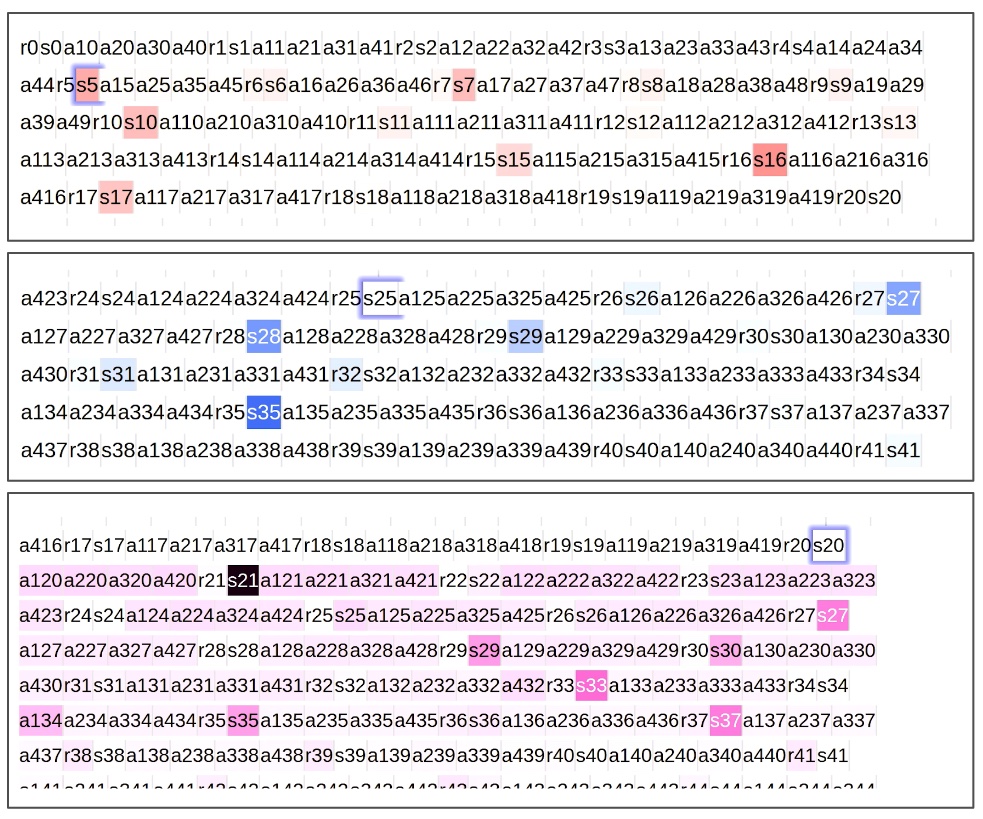}
    \caption{Top to bottom: L3H0, L7H5, L7H7 }
    \label{fig:token-heatmap-2}
\end{figure}

Other interesting patterns observed in the attention head heatmaps are described below. Figure \ref{fig:token-heatmap-2} illustrates another common pattern with a strong focus on certain key states. States s5 (adversary detection), s21 (attack) (also in Figure \ref{fig:token-heatmap-1}, s25 (post-attack) are key states to successfully attack and advance towards the goal state. The focus on key states is a repeated pattern amongst the heads, implying there is some redundancy in what specialisations the heads develop. 

\begin{figure}[h]
    \centering
    \includegraphics[width=1.0\linewidth]{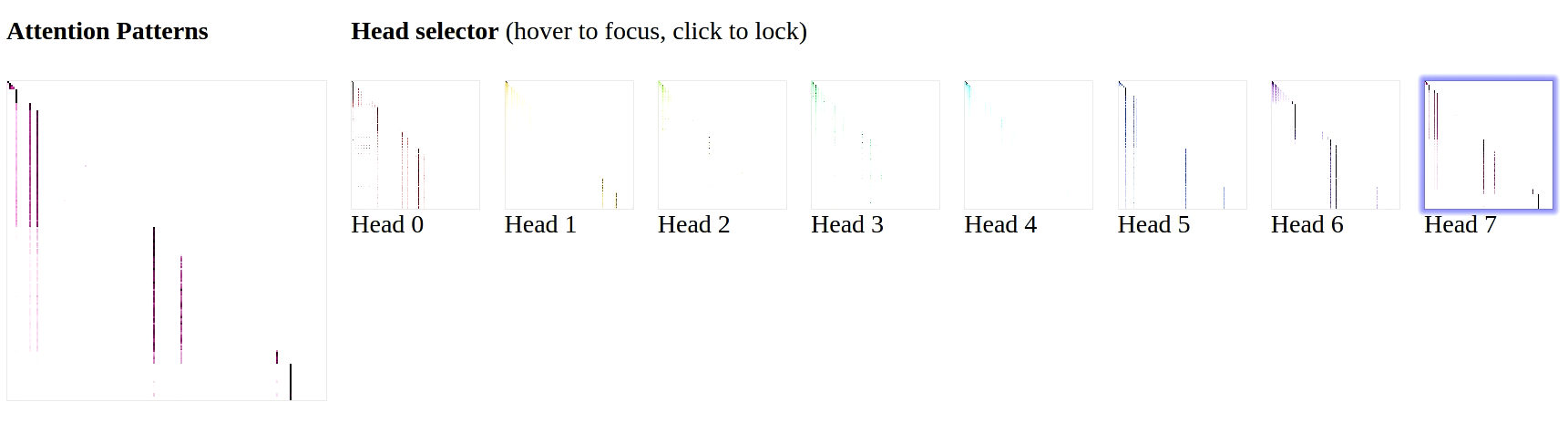}
    \caption{WMH: Attention Layer 2 featuring all 8 heads showing strong focus on key state tokens. }
    \label{fig:mwh-l2}
\end{figure}

\begin{figure}[h]
    \centering
    \includegraphics[width=1.0\linewidth]{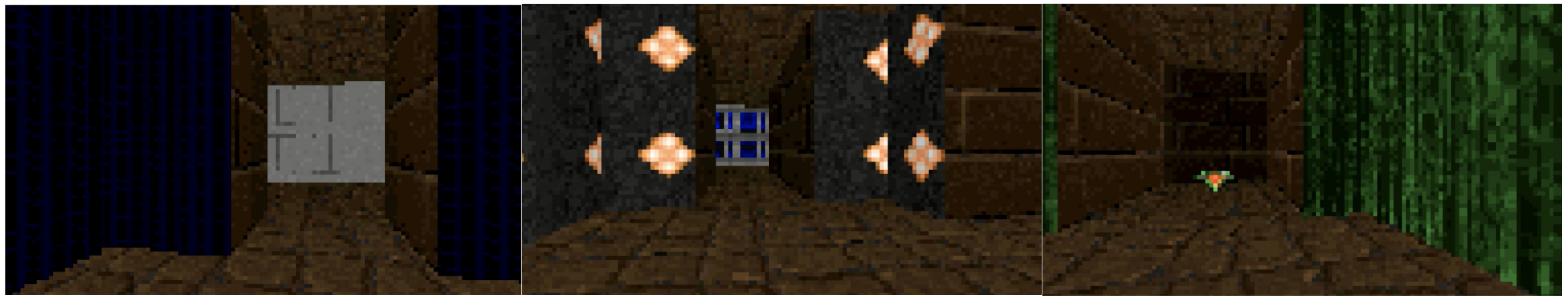}
    \caption{WMH: Key States: s3, s21, s41}
    \label{fig:mwh-key-states}
\end{figure}

Similar patterns focusing on key states emerge in MWH. This evaluation run was 47 steps long and some key states are illustrated in Figure \ref{fig:mwh-key-states}. 
Figure \ref{fig:mwh-l2} focuses on layer 2 where several heads display strong focus on states. Head 7 is particularly interesting and may reflect a handing off from one state to the next in the path towards the goal, s3 to s21 to s41. This pattern repeats in multiple heads in layer 2, most of which focus on states slightly offset from head 7. 

\begin{figure}[h]
    \centering
    \includegraphics[width=1.0\linewidth]{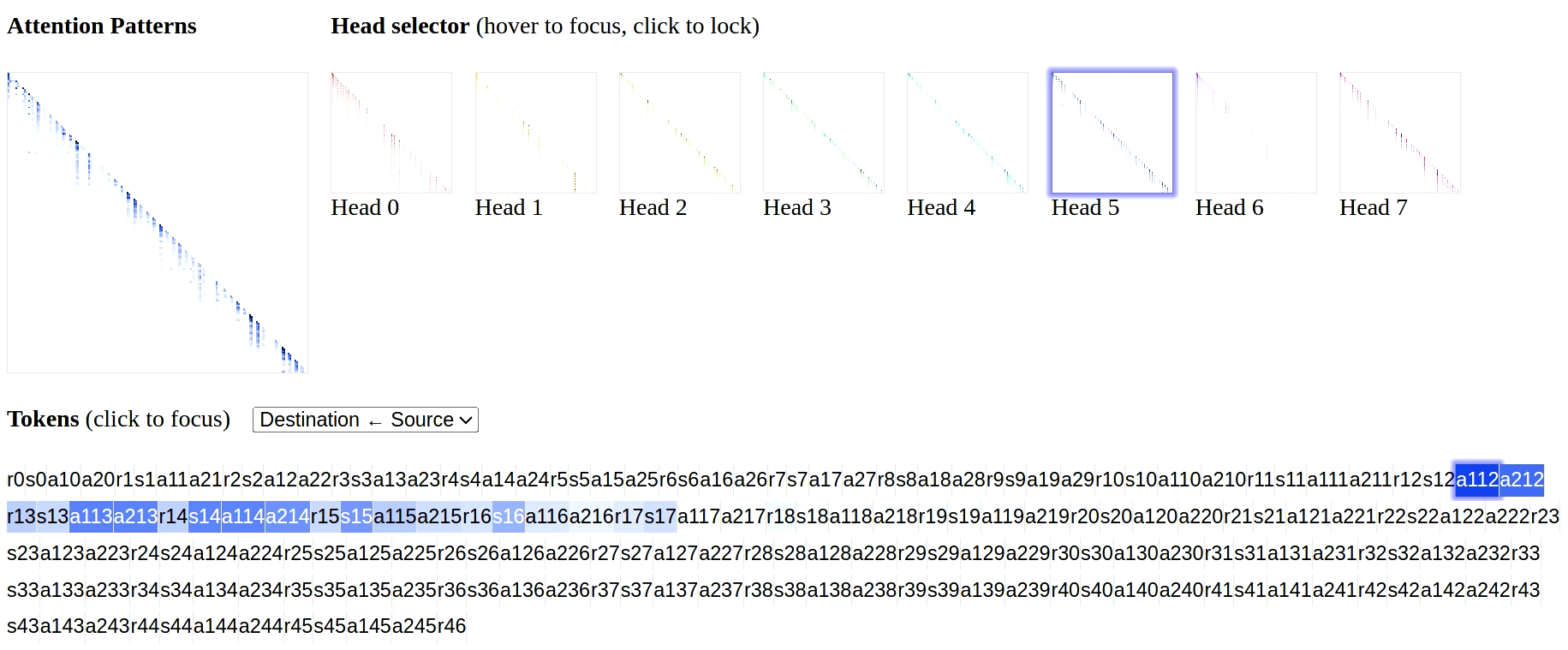}
    \caption{WMH: Layer 6 heads focusing on action tokens}
    \label{fig:mwh-l6-actions}
\end{figure}

\begin{figure}[h]
    \centering
    \includegraphics[width=1.0\linewidth]{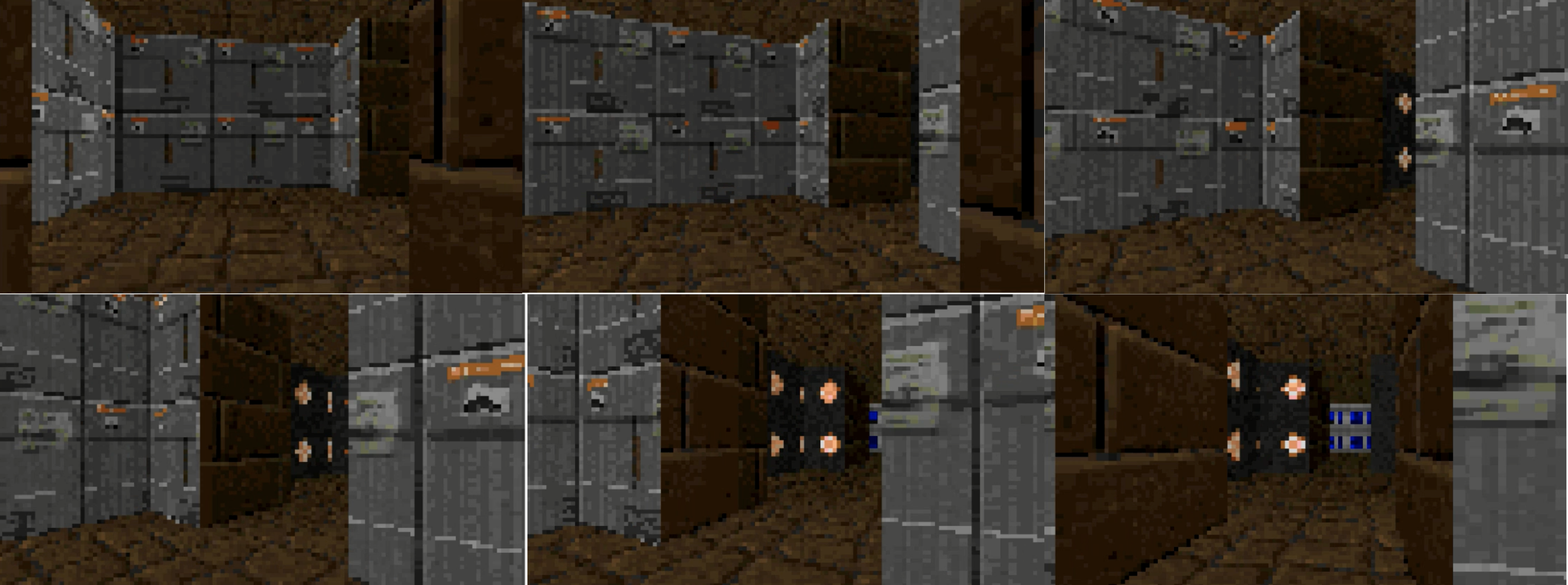}
    \caption{WMH: States 12-17, turning a corner. Read images left to right, top to bottom.}
    \label{fig:mwh-turning}
\end{figure}

Several heads in layer 6 focus on the action tokens. The sequence of tokens influenced by action a112 and a212 illustrated in Figure \ref{fig:mwh-l6-actions} rounds a corner, all steps requiring a similar action pattern for turning. The corresponding states are in Figure \ref{fig:mwh-turning}. 

\clearpage

\end{document}